\documentclass[lettersize,journal]{IEEEtran}
\usepackage{amsmath,amsfonts}
\usepackage{algorithmic}
\usepackage{array}
\usepackage[caption=false,font=normalsize,labelfont=sf,textfont=sf]{subfig}
\usepackage{textcomp}
\usepackage{stfloats}
\usepackage{url}
\usepackage{makecell}
\usepackage{multirow}
\usepackage{booktabs}
\usepackage{verbatim}
\usepackage{graphicx}

\usepackage{amssymb}
\usepackage{multicol}
\usepackage{pifont}
\usepackage{xcolor}
\usepackage{xpatch}
 \usepackage[pagebackref,breaklinks,colorlinks]{hyperref}

\usepackage[capitalize]{cleveref}
\crefname{section}{Sec.}{Secs.}
\Crefname{section}{Section}{Sections}
\Crefname{table}{Table}{Tables}
\crefname{table}{Tab.}{Tabs.}

\usepackage{color}
\usepackage{cite}
\def\BibTeX{{\rm B\kern-.05em{\sc i\kern-.025em b}\kern-.08em
    T\kern-.1667em\lower.7ex\hbox{E}\kern-.125emX}}
\usepackage{balance}

\begin{document}

\title{Knowledge Augmented Relation Inference for Group Activity Recognition}

\author{Xianglong Lang, \and Zhuming Wang, \and
Zun Li, \and
Meng Tian, \and
Ge Shi, \and
Lifang Wu\\
Faculty of Information Technology, Beijing University of Technology, Beijing, China\\
\and
Liang Wang\\
Institute of Automation, Chinese Academy of Sciences, Beijing, China\\
}
\maketitle

\begin{abstract}
   Most existing group activity recognition methods construct spatial-temporal relations merely based on visual representation. Some methods introduce extra knowledge, such as action labels, to build semantic relations and use them to refine the visual presentation. However, the knowledge they explored just stay at the semantic-level, which is insufficient for pursing notable accuracy.
   In this paper, we propose to exploit knowledge concretization for the group activity recognition, and develop a novel Knowledge Augmented Relation Inference framework that can effectively use the concretized knowledge to improve the individual representations. Specifically, the framework consists of a Visual Representation Module to extract individual appearance features, a Knowledge Augmented Semantic Relation Module explore semantic representations of individual actions, and a Knowledge-Semantic-Visual Interaction Module aims to integrate visual and semantic information by the knowledge. Benefiting from these modules, the proposed framework can utilize knowledge to enhance the relation inference process and the individual representations, thus improving the performance of group activity recognition. Experimental results on two public datasets show that the proposed framework achieves competitive performance compared with state-of-the-art methods.
\end{abstract}

\section{Introduction}

Group activity recognition is an important sub-task in the field of video understanding. It shows wide application prospects in intelligent robots, security monitoring, and sports event analysis. 
Unlike the action recognition which focuses on a single individual~\cite{shi2019skeleton,plizzari2021skeleton,c3d,i3d}, group activity recognition needs to understand the scene of multiple individuals. 
This task is more challenging since it relies on the understanding of not only the actions of multiple individuals but also relations among them in the scene. Therefore, both effective individual features and relation modeling are essential to the group activity recognition.

\begin{figure}
    \centering
    \includegraphics[width = 0.97\linewidth]{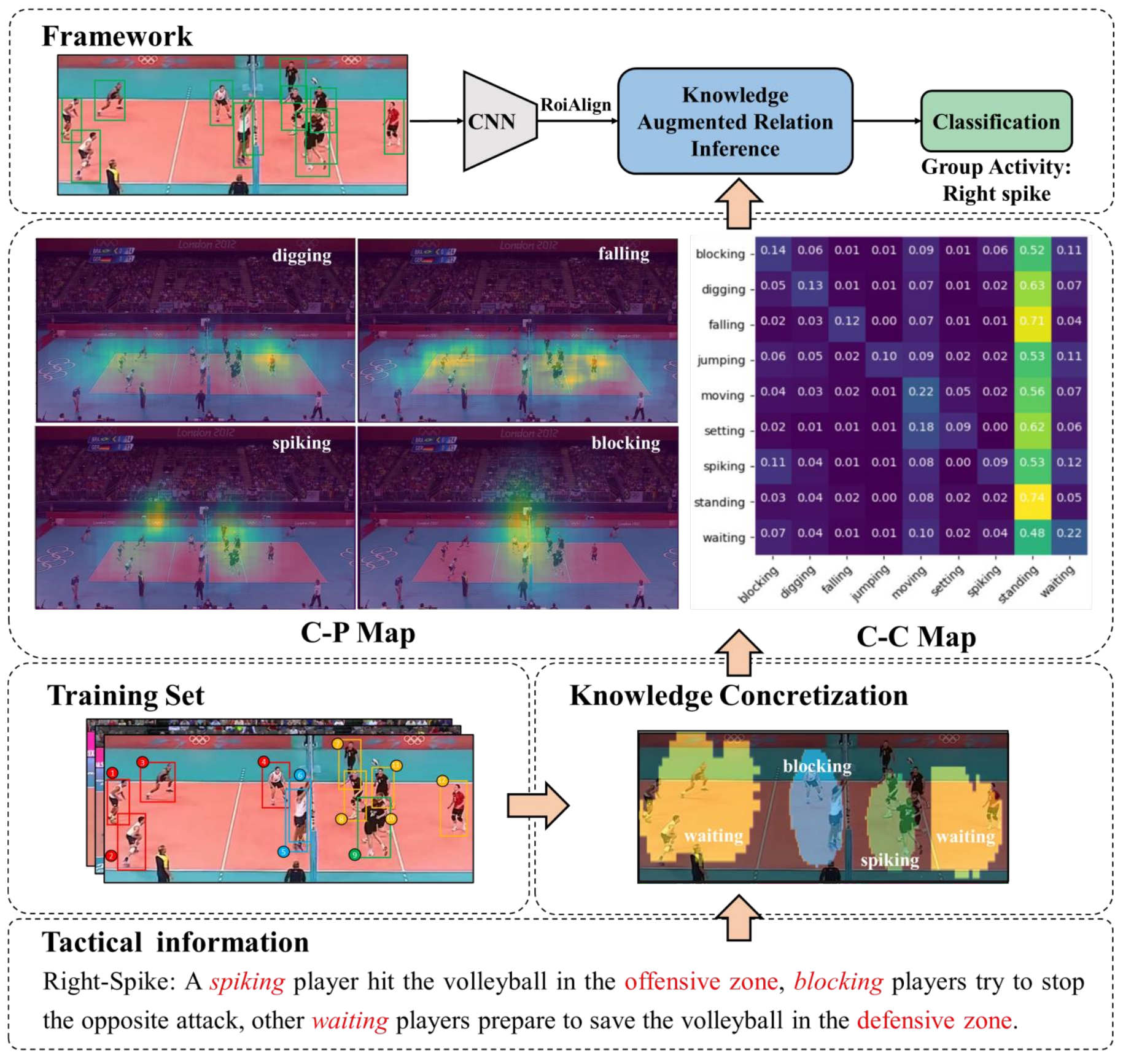}
    \caption{Visualization of our motivation. In team sports, one particular group activity represents the execution and implementation of a specific tactic that reflects the correlation and the distribution of corresponding individual actions. We concretize the abstract knowledge, such as tactics, into the action distribution in different group activities and utilize them to improve the individual representations.}
    \label{fig1}
\end{figure}

Existing methods generally enhance the visual representation of individuals by introducing relation inference~\cite{ibrahim2016hierarchical,bagautdinov2017social,ibrahim2018hierarchical,GLSTM,ARG,gavrilyuk2020actor,yuan2021spatio,li2021groupformer,yuan2021learningcontext,wq,han2022dualai} with the graph network or transformer. However, they build relations mainly based on the visual representations or locations of individuals, which are not completely consistent with the semantic-level individual relations in the group activity. Some methods~\cite{liu2021multimodal,tang2018mining,tang2019learning,contextaware,GINs} introduce extra knowledge, such as action labels, to build semantic relations. By introducing knowledge,
these methods improve the group activity recognition performance well,
but the knowledge they explored merely stay at the semantic-level (\emph{i.e.,} individual action labels), which is insufficient for pursing notable accuracy.

In fact, there is abundant knowledge in real group activity recognition scenarios. For example, in team sports, one particular group activity represents the execution and implementation of a specific tactic that reflects the correlation and distribution of corresponding individual actions. As shown in \cref{fig1}, the \textit{r-spike} activity in volleyball matches involves a ``spiking'' player in the offensive zone, several ``waiting'' players in the defensive zone, and several ``blocking'' players in the opposing offensive zone. Under this description, a \textit{r-spike} activity will never involve a ``setting'' player. This fact provides a clue for distinguishing other activities such as \textit{r-set}. Therefore, if the knowledge can be leveraged more sufficiently, we may have a good chance to improve the reliability of visual representation and interaction modeling, hence further improve the performance of recognition. 

Nevertheless, knowledge is usually extracted from a large amount of samples, it is a kind of highly generalized abstract representation. In contrast, the input samples are concrete. Therefore, it is critical to concretize abstract knowledge into the same space as input samples. 
Although several methods utilize  cross-model aggregation~\cite{liu2021multimodal} or knowledge distillation~\cite{tang2019learning} to concretize the semantic label as the latent feature vector for the visual representation interacting, such concretization manner is hard to leverage richer knowledge. 
In fact, group activity is a comprehensive expression of a group of individual actions, which is correlated with individual actions and their position, 
and such correlation is implied in a large number of samples. Thus, it is possible to obtain a concrete representation of knowledge from amount of training samples through statistics.

In this paper, we propose to concretize the abstract knowledge, such as tactics, into the action distribution in different group activities, which is further represented as Class-Class Distribution Map (C-C Map) and Class-Position Distribution Map (C-P Map). They present the correlation and distribution of individual actions. Furthermore, we propose a novel Knowledge Augmented Relation Inference Framework to construct interactions among individuals and use the above two maps to enhance the individual representations for group activity recognition. Specifically, we first design a Visual Representation Module to extract the individual appearance representations. Then we design a Semantic Relation Module to construct the correlation between different individual actions with the assistance of the C-C Map. After that, a Knowledge-Semantic-Visual Interaction Module is devised to integrate visual information and semantic information through a cross-modal interacting block, combining the C-P Map to perform the relation inference and improve the individual representation ability. Finally, the enhanced individual features, along with raw visual features, are utilized for activity recognition. We evaluate our method on the Volleyball dataset and the Collective Activity dataset, and the experimental results show that the proposed framework achieves competitive performance compared to the state-of-the-art methods.

The contributions of this paper are summarized below:
\begin{itemize}
\item We propose an idea of knowledge concretization for the group activity recognition. And the knowledge in the specific application scenarios such as team sports or surveillance is concretized as the Class-Class Distribution Map (C-C Map) or Class-Position Distribution Map (C-P map). 

\item We propose a novel Knowledge Augmented Relation Inference framework which integrates visual representation and knowledge (\emph{i.e.,} action labels, C-C Map, C-P Map) in a unified relation inference architecture. 

\item Experiments on two public datasets show that the proposed method achieves competitive results compared with state-of-the-art methods. And the introduction of knowledge is also helpful in improving performance with limited training data.
\end{itemize}

\section{Related Work}
Group activity recognition has been studied for over a decade. A lot of methods have been proposed. Early methods extract hand-crafted features to infer group behavior by probabilistic graphical models~\cite{amer2012cost,lan2012social,amer2014hirf,choi2013understanding,Amer2013}. Recently, deep relation inference based methods have shown promising performance~\cite{wu2021comprehensive}. They can be generally classified into the visual representation based method and visual-semantic representation based methods, according to the information they used.

\begin{figure*} 
    \centering
    \includegraphics[width=0.95\linewidth]{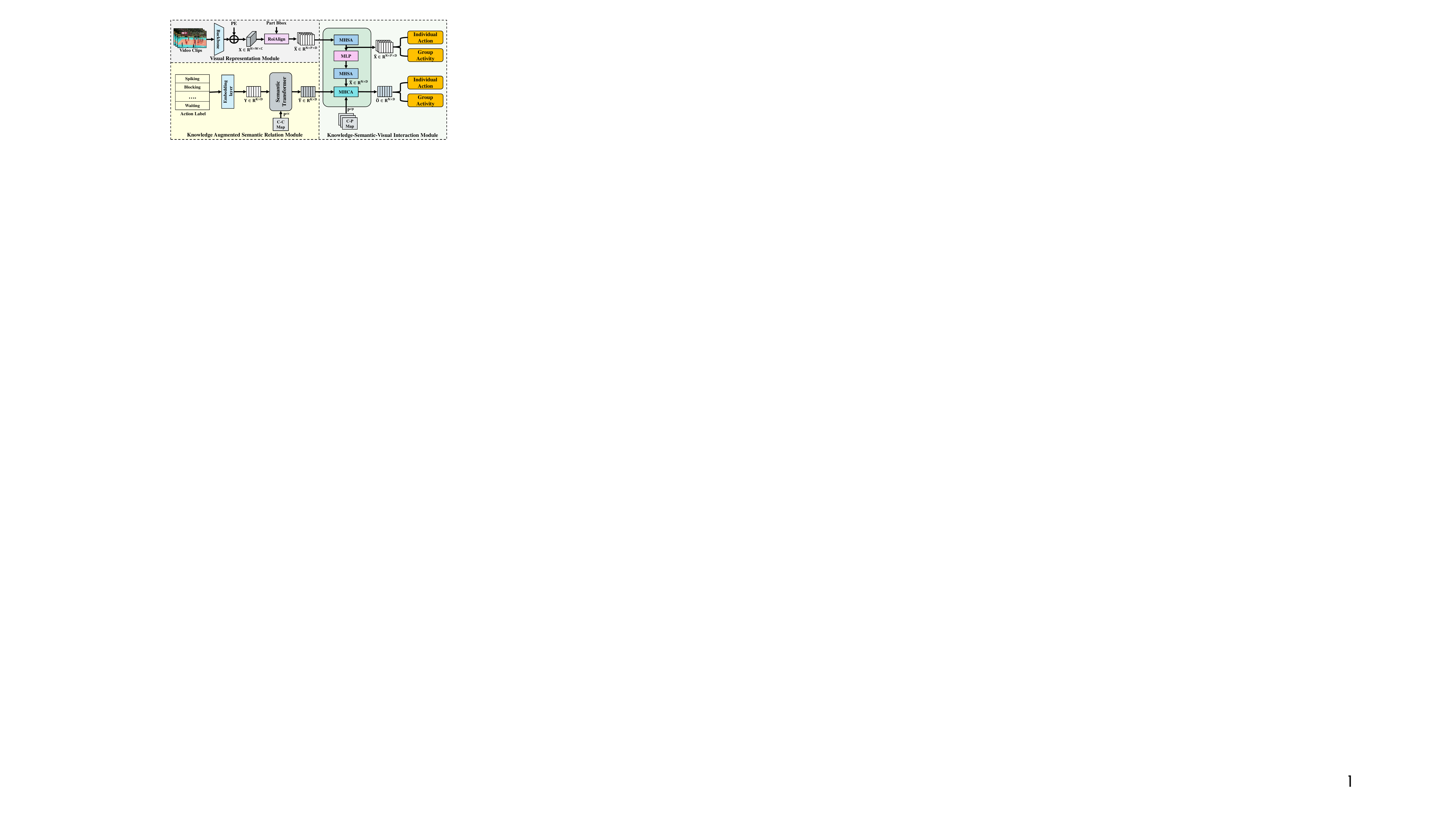}
    \caption{Overview of the proposed framework. It first devises the Visual Representation Module to extract the appearance representation of individuals. And then, it design a Knowledge Augmented Semantic Relation Module to capture semantic representations of individual actions with the assistance of the Class-Class Distribution Map. After that, a Knowledge-Semantic-Visual Interaction Module is proposed to integrate visual and semantic information, and enhance individual representations with the help of the Class-Position Distribution Map. Finally, the individual representations are utilized to perform the final classification.} 
    \label{fig2}
\end{figure*} 

\textbf{Visual Representation based Methods.}  
Visual representation based methods usually obtain the enhanced visual representation by introducing visual relation inference~\cite{ibrahim2016hierarchical,wang2017recurrent,bagautdinov2017social,ibrahim2018hierarchical,stagNet,tang2019learning,azar2019convolutional,hu2020progressive,yuan2021learningcontext,Detector-Free,han2022dualai}. Some methods adopt RNN or LSTM to explore the individual spatio-temporal relation in scene~\cite{ibrahim2016hierarchical,shu2017cern,bagautdinov2017social,stagNet,HANHCN,CCG,yan2018participation,PCTDM}. Alternatively, some researchers introduce attention mechanism to relation inference~\cite{ARG,yan2020higcin,lu2019gaim,yuan2021spatio,gavrilyuk2020actor,ehsanpour2020joint,li2021groupformer,Detector-Free, han2022dualai}, and improve the representation of visual features. Wu {\it et al}.~\cite{ARG} utilize a graph structure to construct the relations between actors in the scene and enhance their representation by graph convolution network. Yan {\it et al}.~\cite{yan2020higcin} construct a cross-graph to explore the temporal dynamics and spatial interaction context. Yuan {\it et al}.~\cite{yuan2021spatio} use the well-designed dynamic relation module and dynamic walk module to build person-specific interaction, which can model spatial-temporal effectively. Gavrilyuk {\it et al}.~\cite{gavrilyuk2020actor} adopt transformer architecture to model spatial-temporal relations among individuals and use a multi-modal information fusion strategy. Li {\it et al}.~\cite{li2021groupformer} propose a Clustered Spatial-Temporal Transformer to deeply explore the correlation of spatial and temporal context in a parallel manner. Yuan and Ni~\cite{yuan2021learningcontext} encode the global contextual information into individual features and explore all pairwise interactions between individuals. Han {\it et al}.~\cite{han2022dualai} propose a Dual-path Actor Interaction framework to learn complex actor relations in videos and further enhance individual representation by using an efficient self-supervised signal.

\textbf{Visual-Semantic Representation based Methods.} 
Visual-semantic representation based methods introduce semantic information to relation inference and improve the consistency of visual relations with the semantic-level individual relations in the group activity~\cite{sbgar,stagNet,tang2018mining,GINs,contextaware,liu2021multimodal}. Li {\it et al}.~\cite{sbgar} propose a novel semantics based scheme that recognizes group activities based on the semantic meaning of video captions generated by LSTM. Qi {\it et al}.~\cite{stagNet} introduce a semantic graph to explicitly describe the spatial content of the scene and employ a structural-RNN to incorporate it with the temporal factor. Liu {\it et al}.~\cite{liu2021multimodal} directly utilize individual action labels to construct a semantic graph to refine visual representations. Tang {\it et al}.~\cite{tang2018mining} adopt knowledge distillation to force individual visual representations to be consistent with semantic representations embedded from action labels. 

The existing methods demonstrate that the introduction of extra knowledge is helpful in improving visual representation. However, the knowledge they explored is merely a small part of knowledge in the real application scenarios, and the way of knowledge utilization can not adapt to the complicated knowledge. Unlike these methods, we utilize richer knowledge, such as tactical information, and introduce the concretized knowledge into the relation inference framework to improve the feature representation.

\section{Method}
\subsection{Overall Architecture}
\label{sec3.1}
Our framework is mainly composed of three modules: the Visual Representation Module for extracting appearance representation of individuals, the Knowledge Augmented Semantic Relation Module for encoding semantic representations of individual actions, and the Knowledge-Semantic-Visual Interaction Module for  aggregating visual and semantic information. As illustrated in \cref{fig2}, our framework first summarizes the individual actions, and constructs a Class-Class Distribution Map (C-C Map) and a Class-Position Distribution Map (C-P Map). Then, it feds C-C Map into the Knowledge Augmented Semantic Relation Module to enhance the semantic representation. Afterward, it utilizes the semantic representation and the C-P Map to enhance the visual representation through the Knowledge-Semantic-Visual Interaction Module. 
Finally, it predicts the group activities using the enhanced visual representations.

\subsection{Knowledge Concretization}
\label{sec3.2}

As discussed above, knowledge is an abstract semantic representation with a gap from training samples. Thus, prior to the training procedure, we concretize knowledge in a form that can be integrated with visual representation of samples. To be specific, we summarize the individual actions of samples in the training set to concretize the semantic representation of knowledge, and further construct a Class-Class Distribution Map (C-C Map) and a Class-Position Distribution Map (C-P Map). These two maps can present the correlation and distribution of individual actions in one particular group activity.

\textbf{Class-Class Distribution Map.} 
Given $K$ different individual action labels ${\mathbf{L}}=\{{l_i}\}_{i=1}^K$ in the training set, we count the total concurring times $m_{ij}$ of the $i$-th action label $l_{i}$ and the $j$-th action label $l_j$. For example, if two ``blocking'' players and three ``standing'' players occur in an image, we record the concurring times of ``blocking'' and ``standing'' in this image as 6. Then we add up their concurring times in all images to get the total concurring times of ``blocking'' and ``standing''. In this way, we construct the Class-Class Distribution Map $\mathbf{P^{cc}}\in {\mathbb{R}^{K\times K}}$, which measures the correlation degree among individual action labels, as follow:
\begin{equation}
    p^{cc}_{ij}=\frac{{{m}_{ij}}}{\sum\limits_{i=1}^K\sum\limits_{j=1}^K{m_{ij}}}
\end{equation}
where $p^{cc}_{ij}\in\mathbf{P^{cc}}$ corresponds to the correlation of $i$-th and $j$-th label. 
This value reflects the probability of the simultaneous occurrence of different actions in a specific scenario.

\textbf{Class-Position Distribution Map.}
On the middle frame of every video clip in the training set, for the $i$-th individual action, we mark the coordinate $(x,y)$ of each individual who performs it. Then we project these coordinates of all video clips onto one single image $b_{i}$, which shares the same size with input frames. This way, we can obtain the distribution maps of $K$ individual action labels. Similar to the Class-Class Distribution Map, we construct the Class-Position Distribution Map $\mathbf{P^{cp}}\in {\mathbb{R}^{H\times W\times K}}$, which represents the distribution of individual actions, as follow:
\begin{equation}
    p^{cp}_{ixy}=\frac{{b_{ixy}}}{\sum\limits_{x=1}^H\sum\limits_{y=1}^Wb_{ixy}}
\end{equation}
where $p^{cp}_{ixy}\in\mathbf{P^{cp}}$ denotes the value of $\mathbf{P^{cp}}$ of the $i$-th individual action at the coordinate of $(x,y)$.
And $b_{ixy}$ denotes the value of $b_{i}$ at the coordinate of $(x,y)$, it reflects the occurrence probability of each individual action in a specific spatial location. 

\subsection{Visual Representation Module}
\label{sec3.3}
The Visual Representation Module aims to extract the appearance features of individuals. As shown in \cref{fig2}, given a $T$ frames video clip, we adopt an inflated 3D convNets network (I3D)~\cite{i3d} pre-trained on Kinetics dataset~\cite{kay2017kinetics} as the backbone to extract image appearance features, and employ two dimensions positional encoding (PE) to provide position information as in \cite{detr,wang2021end}. In this way, we obtain the raw individual visual representation $\mathbf{X} \in \mathbb{R}^{H \times W  \times C}$, where $C$ is the number of channels. $\mathbf{X}$ also represents the global scene information directly extracted from the input frame. Then, the RoiAlign~\cite{he2017mask} operation is applied to extract the refined visual features of individuals from $\mathbf{X}$ according to the body part bounding boxes of each individual in the scene. After that, we utilize the fully-connected layer and ReLU~\cite{nair2010rectified} activation function to encode the extracted features as a $D$ dimensional feature vector $\mathbf{\overline{X}} \in \mathbb{R}^{N \times P  \times D}$, where $N$ represents the number of actors in the scene, $P$ is the number of body parts. $\mathbf{\overline{X}}$ presents the visual representation of individuals and is further utilized to perform the individual relational inference in a later module.

\begin{figure}
  \centering
   \includegraphics[width=0.83\linewidth]{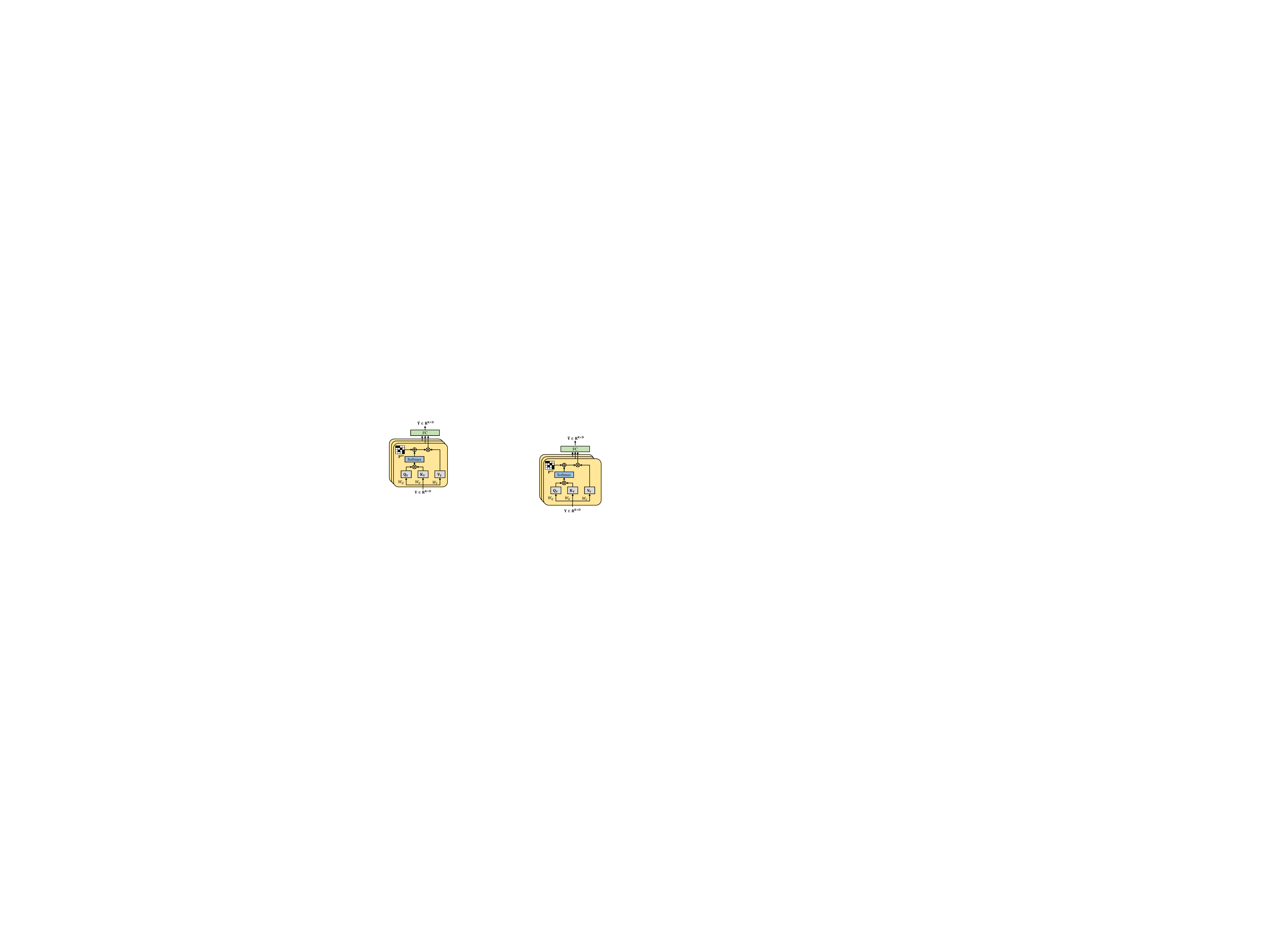}
   \caption{Illustration of multi-head self-attention in the Semantic Transformer. This process aims to explore the correlation between different action labels, and introduces C-C Map by element-wise operation to impact semantic relation modeling process.}
   \label{SRM}
\end{figure}

\subsection{Knowledge Augmented Semantic Relation Module}
\label{sec3.4}
In this module, we first encode $K$ different individual action labels $\mathbf{L}$ into one-hot vectors and embed them into a $D$ dimension latent space to obtain the semantic features $\mathbf{Y} \in {\mathbb{R}^{K\times D}}$. After that, we propose a Semantic Transformer to infer the relation among semantic features of different individual actions, and introduce the C-C Map into the multi-head self-attention mechanism of the standard transformer. In conventional self-attention operation, the output is computed as a weighted sum of values ($V$), where the weights are computed by the correlation function of queries ($Q$) and keys ($K$). As shown in \cref{SRM}, to better explore the correlation of different semantic features, we add C-C Map to the weights of values before the weighted sum operation. In this way, we can use real data distribution to facilitate the relation modeling.
The output of multi-head self-attention mechanism $\mathbf{\overline{Y}}$ can be formulated as:
\begin{equation}
    A^s_i=\sigma\left(\frac{\mathbf{Y}W^Q_i\cdot{(\mathbf{Y}W^K_i)}^T}{\sqrt{d}}\right)
\end{equation}
\begin{equation}
    h^s_i = \left(A^s_i+\mathbf{P^{cc}}\right)\cdot\mathbf{Y}W^V_i
\end{equation}
\begin{equation}
    \mathbf{\overline{Y}}=\mathrm{F}^s([{h}^s_1,{h}^s_2,...,{h}^s_i])
\end{equation}
where $W^Q_i,W^K_i,W^V_i$ are the learnable matrices whose dimension is $D \times d$, $i$ is the number of attention heads. $\sigma$ presents the softmax operation. $[,]$ denotes the concatenate operation. $\mathrm{F}^s$ refers to the fully-connected layer adopted to integrate the outputs of multiple attention heads.

The residual connection operation and a feed-forward network are adopted to enhance the feature representation, and the final output of the Semantic Transformer is denoted as 
$\mathbf{\widehat{Y}}\in \mathbb{R}^{K \times D}$.

\begin{figure}
  \centering
   \includegraphics[width=0.97\linewidth]{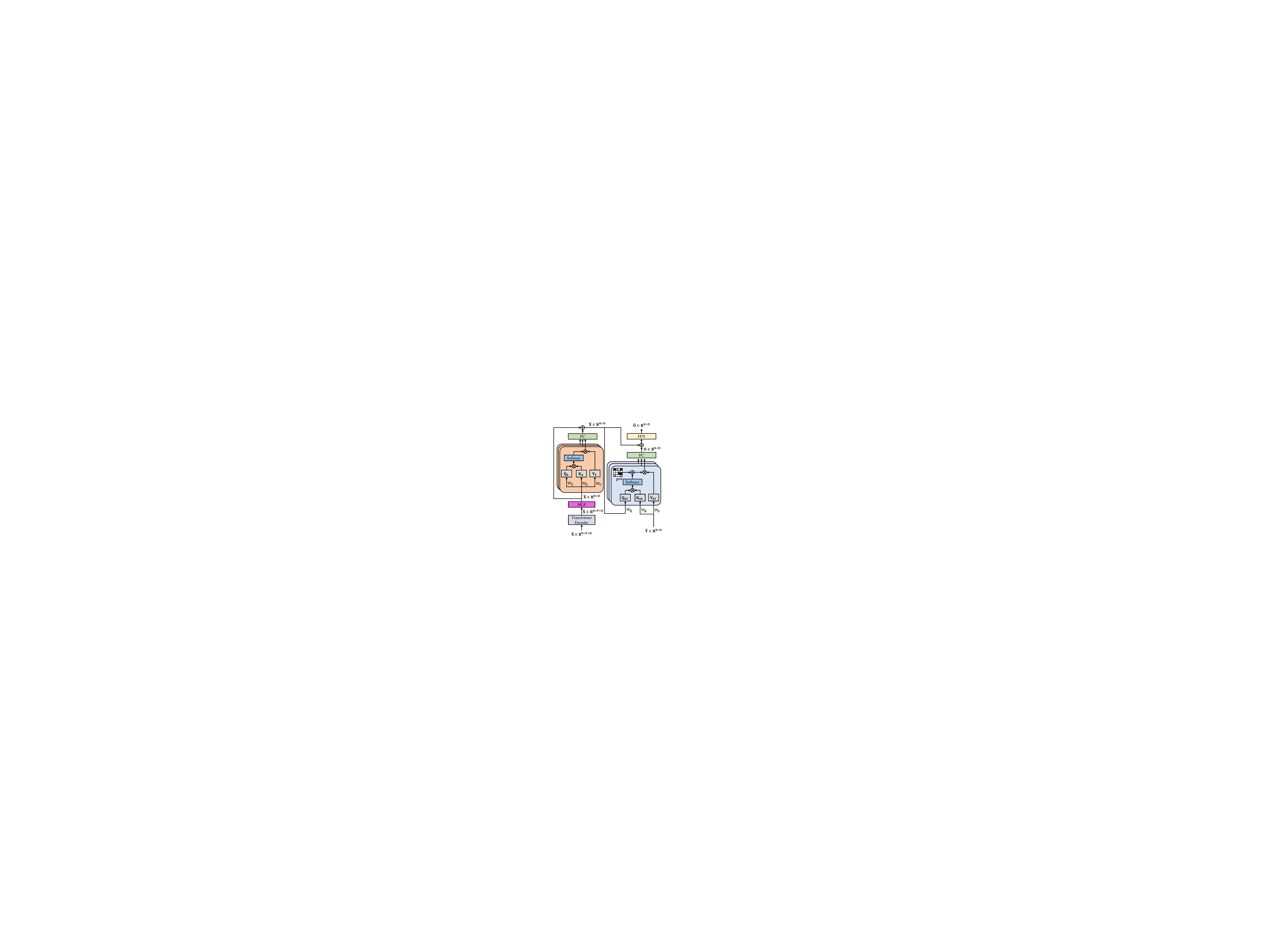}
   \caption{Illustration of the Knowledge-Semantic-Visual Interaction Module. This module integrates visual and semantic information to enhance the individual representations, and introduces C-P Map into the process of multi-head cross attention mechanism.}
   \label{HRM}
\end{figure}

\subsection{Knowledge-Semantic-Visual Interaction Module}
\label{sec3.5}
This module enhances the individual representations by integrating visual and semantic representation with the assistance of the C-P Map, and performs the task of group activity recognition. We first employ a conventional vision transformer encoder~\cite{gavrilyuk2020actor} to enhance $\mathbf{\overline{X}}$ output from Visual Representation Module and obtain a refined individual representation $\mathbf{\widehat{X}} \in \mathbb{R}^{N \times P \times D}$ which performed preliminary relational inference, and a multilayer perceptron to adjust the shape of $\mathbf{\widehat{X}}$ into ${N \times D}$. Then, we design a Visual-Semantic Inference Transformer to perform individual relation inference across the visual and semantic representation. Specifically, we feed $\mathbf{\widehat{X}}$ into the encoder of the transformer and output the encoded visual representation. The multi-head self-attention mechanism used in the above encoder is similar to which in the Semantic Transformer we introduced in \cref{sec3.4}, yet the C-C Map is not added before the weighted sum operation. The sum of $\mathbf{\widehat{X}}$ and the output of the self-attention mechanism is denoted as $\mathbf{\widetilde{X}}\in \mathbb{R}^{N \times D}$ and further used in the subsequent process. 

The decoder of the Visual-Semantic Inference Transformer takes the enhanced semantic feature representation $\mathbf{\widehat{Y}}$, the encoded visual representation $\mathbf{\widetilde{X}}$ and C-P Map as input to export the knowledge augmented individual features $\mathbf{\overline{O}} \in {\mathbb{R}^{N\times D}}$. In addition, to better correspond the individuals with the distribution of their actions, we first select the corresponding $N$ position coordinates from the C-P Map according to the bounding box of $N$ individuals on the frame and convert the dimension of the C-P Map into $N\times K$ from $H\times W\times K$. As shown in \cref{HRM}, we use a multi-head cross-attention mechanism to perform semantic-visual interaction as follow: 
\begin{equation}
    {A}^o_i=\sigma\left(\frac{\mathbf{\widetilde{X}}{W^{\widehat{Q}}_i}\cdot{(\mathbf{\widehat{Y}}{W^{\widehat{K}}_i})}^T}{\sqrt{d}}\right)
\end{equation}
\begin{equation}
    {h}^o_i = \left(A^o_i+\mathbf{P^{cp}}\right)\cdot\mathbf{\widehat{Y}}{W^{\widehat{V}}_i}
\end{equation}
\begin{equation}
    \mathbf{O}=\mathrm{F}^o([{h}^o_1,{h}^o_2,...,{h}^o_i])
\end{equation}
where ${W^{\widehat{Q}}_i},{W^{\widehat{K}}_i},{W^{\widehat{V}}_i}$ are the learnable matrices whose dimension is $D \times d$, $i$ is the number of attention heads. $\mathrm{F}^o$ denotes the fully-connected layer adopted to integrate the outputs of multiple attention heads. In this way, we can take the distribution of individual actions to guide the representation of features.

The residual connection operation and a feed-forward network are adopted to enhance the feature representation, and output the knowledge augmented individual feature $\mathbf{\overline{O}}$.

\subsection{Training and Reasoning}
\label{sec3.6}
Our framework is trained in an end-to-end manner. To supervise the feature representation learning, we design two classification heads to perform the individual action classification and group activity classification for both $\mathbf{\widehat{X}}$ and $\mathbf{\overline{O}}$. We also utilize global scene information $\mathbf{X}$ to perform group activity classification. The classification losses $\mathcal{L}_{x}$, $\mathcal{L}_{o}$ and $\mathcal{L}_{s}$ can be formulated as: 
\begin{equation}
    \mathcal{L}_{x} =  \mathcal{L}_{CE}(\mathbf{\hat{y}}^{x}_\mathbf{g},{\mathbf{y}_\mathbf{g}})+\lambda \mathcal{L}_{CE}(\mathbf{\hat{y}}^{x}_{a},{\mathbf{y}_{a}})
\end{equation}
\begin{equation}
    \mathcal{L}_{o} =  \mathcal{L}_{CE}(\mathbf{\hat{y}}^{o}_\mathbf{g},{\mathbf{y}_\mathbf{g}})+\lambda \mathcal{L}_{CE}(\mathbf{\hat{y}}^{o}_{a},{\mathbf{y}_{a}})
\end{equation}
\begin{equation}
    \mathcal{L}_{s} =  \mathcal{L}_{CE}(\mathbf{\hat{y}}^{s}_\mathbf{g}, {\mathbf{y}_\mathbf{g}})
\end{equation}
where $\mathcal{L}_{CE}(\cdot)$ denotes the cross-entropy loss function. $\mathbf{y}_{a}$, $\mathbf{y}_\mathbf{g}$ are the ground truth labels for individual actions and group activities, receptively. $\mathbf{\hat{y}}^{x}_{a}$ and $\mathbf{\hat{y}}^{o}_{a}$ represent individual action scores predicted from $\mathbf{\widehat{X}}$ and $\mathbf{\overline{O}}$. Similarly, $\mathbf{\hat{y}}^{x}_\mathbf{g}$ and $\mathbf{\hat{y}}^{o}_\mathbf{g}$ represent group activity scores predicted from $\mathbf{\widehat{X}}$ and $\mathbf{\overline{O}}$, respectively.
 $\mathbf{\hat{y}}^{s}_\mathbf{g}$ is group activities scores predicted from $\mathbf{X}$. $\lambda$ is the scalar weight to balance different classification tasks. The overall loss function is formed as follow:
\begin{equation}
    \mathcal{L} =  \mathcal{L}_{x} + \mathcal{L}_{o} + \mathcal{L}_{s}
\end{equation}

In the inference stage, we sum the group activity classification scores $\mathbf{\hat{y}}^{x}_\mathbf{g}$, $\mathbf{\hat{y}}^{o}_\mathbf{g}$ and $\mathbf{\hat{y}}^{s}_\mathbf{g}$ as the final classification score.

\section{Experiments}
\subsection{Datasets}
\label{sec4.1}
\textbf{Volleyball dataset.}
The Volleyball dataset (VD) is one of the largest public datasets for evaluation of the group activity recognition.
 It has 3,493 and 1,337 video clips for training and testing, respectively. 
Moreover, it provides high-resolution video clips containing eight group activity categories, including left spike, right spike, left set, right set, left set, right set, left set, right set, left win, and right win. In the middle frame of each video clip, all individuals are labeled by bounding box coordinates and individual action categories (blocking, digging, falling, jumping, moving, setting, passing, spiking, standing, and waiting). 
Image resolution for VD is 1280 $\times$ 720.

\textbf{Collective Activity dataset.}
The Collective Activity dataset (CAD) is another high-quality public dataset for group activity recognition. It contains 5 group activity categories (walking, crossing, waiting, talking, and queuing). The middle frame of every ten frames in this dataset is labeled with bounding box coordinates and individual action categories (NA, walking, crossing, waiting, talking, and queuing). The group activity category is determined by the vast majority of individual categories in the scene. Image resolution for CAD is 720 $\times$ 480.

\subsection{Implementation Details}
\label{sec4.2}
For each video clip, we select ten frames (middle frame, five frames before, and four frames after) as the input of the backbone network. We utilize the RoIAlign layer with crop size $7 \times 7$ to obtain $\mathbf{\overline{X}}$ and embed them into $D=256$. For experiments on VD, we employ two attention heads in the encoder layer of transformers and one attention head in the decoder. And for experiments on CAD, we set the number of attention heads in the encoder of transformers to 16. The dimension of $d$ is set to 128 and the size of the fully-connected layer in the feed-forward network is set to 1024. For the VD, we utilize Adam optimizer with $\beta_{1}=0.9, \beta_{2}=0.999$ and $\varepsilon=10^{-8}$, empirically. The batch size is set to 1, the learning rate ranging from $1 \times 10^{-4}$ to $1 \times 10^{-6}$. For the CAD, we set Adam optimizer hyper-parameters $\varepsilon=10^{-10}$, the batch size to 2, and the initial learning rate as $5 \times 10^{-5}$. The other settings are the same as on the Volleyball dataset. We adopt the widely used Multi-class Classification Accuracy (MCA) and Mean Per Class Accuracy (MPCA) as evaluation metrics. Our experiments are conducted on an NVIDIA GeForce GTX 2080 GPU with PyTorch deep learning framework.

\begin{table}
  \centering
  \caption{Comparison with state-of-the-art methods on the Volleyball dataset.}
  \begin{tabular}{@{}lccc@{}}
    \toprule
    Method & Backbone & \makecell{Optical \\ Flow} & MCA \\
    \midrule
        HDTM~\cite{ibrahim2016hierarchical} & AlexNet & & 81.9\\
        CERN~\cite{shu2017cern} & Vgg16 & & 83.3\\
        StagNet~\cite{stagNet} & Vgg16 & & 89.3\\
        Detector-Free~\cite{Detector-Free} & Resnet-18 & & 90.5\\
        SSU~\cite{bagautdinov2017social} & Inception-v3 & & 90.6\\
        HiGCIN~\cite{yan2020higcin} & Resnet-18 & & 91.4\\
        AT~\cite{gavrilyuk2020actor} & I3D & & 91.4\\
        PRL~\cite{hu2020progressive} & Vgg16 & & 91.4\\
        ARG~\cite{ARG} & Inception-v3 & & 92.5\\
        STBiP~\cite{yuan2021learningcontext} & Inception-v3 & & 93.3\\
        STDIN~\cite{yuan2021spatio} & Vgg16 &  & 93.6\\
        Groupformer~\cite{li2021groupformer} & Inception-v3 & & 94.1\\
        Dual-AI~\cite{fu2019dual} & Inception-v3 & & 94.4\\
        \hline
        SBGAR~\cite{sbgar} & Inception-v3 & \checkmark & 66.9\\
        CRM~\cite{azar2019convolutional} & I3D & \checkmark & 93.0\\
        AT~\cite{gavrilyuk2020actor} & I3D & \checkmark & 93.0\\
        JLSG~\cite{ehsanpour2020joint} & I3D   & \checkmark & 93.1\\
        MSCA-GNN~\cite{liu2021multimodal} & I3D  & \checkmark & 93.4\\
        ERN~\cite{pramono2020empowering} & R50-FPN+I3D & \checkmark & 94.1\\
        Groupformer~\cite{li2021groupformer} & I3D & \checkmark & 94.9\\
        Dual-AI~\cite{han2022dualai} & Inception-v3 & \checkmark & \textbf{95.4}\\
    \hline
        Ours(RGB) & I3D & & \textbf{94.5}\\
        Ours(RGB+Flow) & I3D & \checkmark & 94.8\\
    \bottomrule
  \end{tabular}
  \label{table1}
\end{table}

\begin{table}[h]
  \centering
  \caption{Comparison against state-of-the-art methods on the Volleyball dataset under limited training data.}
  \begin{tabular}{@{}lccccc@{}}
    \toprule
    \multirow{2}{*}{Method} & \multicolumn{5}{c}{Data Ratio} \\
    \cmidrule{2-6}
    & 5\% & 10\% & 25\% & 50\% & 100\% \\
    \midrule
        PCTDM~\cite{PCTDM} & 53.6 & 67.4 & 81.5 & 88.5 & 90.3 \\
        AT~\cite{gavrilyuk2020actor} & 54.8 & 67.7 & 84.2 & 88.0 & 90.0 \\
        HiGCIN~\cite{yan2020higcin} & 35.5 & 55.5 & 71.2 & 79.7 & 91.4\\
        ERN~\cite{ehsanpour2020joint} & 41.2 & 52.5 & 73.1 & 75.4 & 90.7\\
        ARG~\cite{ARG} & 69.4 & 80.2 & 87.9 & 90.1 & 92.3\\
        STDIN~\cite{yuan2021spatio} & 58.3 & 71.7 & 84.1 &  89.9 & 93.1\\
        Dual-AI~\cite{han2022dualai} & 76.2 & 85.5 & 89.7  & 92.7 & 94.4\\
    \hline
        Ours-Base & 66.2 & 78.8 & 88.2 & 92.2 & 93.4\\
        \textbf{Ours} & \textbf{79.0} & \textbf{85.6} & \textbf{92.1}  & \textbf{93.2} & \textbf{94.5}\\
    \bottomrule
  \end{tabular}
  \label{table2}
\end{table}

\begin{table}
  \centering
  \caption{Comparison with state-of-the-art methods on the Collective Activity dataset.}
  \begin{tabular}{@{}lccc@{}}
    \toprule
    Method & Backbone  & MCA & MPCA\\
    \midrule
    SBGAR~\cite{sbgar} & Inception-v3  & 86.1 & -\\
    Recurrent~\cite{wang2017recurrent} & Vgg16 & - & 89.4\\
    PCTDM~\cite{yan2018participation} &	AlexNet & - & 92.2\\
    PRL~\cite{hu2020progressive} & Vgg16 & - & 93.8\\
    CRM~\cite{azar2019convolutional} & I3D  & 85.8 & 94.2\\
    JLSG~\cite{ehsanpour2020joint} & I3D   & 89.4 & -\\
    SPTS \cite{tang2018mining} & Vgg16 &	90.7 & 95.7\\
    AT~\cite{gavrilyuk2020actor} & I3D & 92.8 & 98.5\\
    MSCA-GNN~\cite{liu2021multimodal} & I3D   & 93.1 & - \\
    ERN~\cite{pramono2020empowering} & R50-FPN+I3D &  93.9 & - \\
    Groupformer~\cite{li2021groupformer} & I3D & \textbf{94.7} & -\\
    Dual-AI~\cite{han2022dualai} & Inception-v3  & - & 96.5\\
    \hline
    Ours(RGB) & I3D & 92.8 & 98.5 \\
    Ours(RGB+Flow) & I3D & 93.5 & \textbf{98.7}\\
    \bottomrule
  \end{tabular}
  \label{table3}
\end{table}

\subsection{Comparison with the State-of-the-Arts}
\label{sec4.3}
\textbf{Results on Volleyball dataset.}
We compare our framework with the state-of-the-art methods on VD and report the results in \cref{table1}. Our method (RGB only) reaches the MCA of 94.5\%, achieving the best performance among all comparison methods~\cite{ibrahim2016hierarchical,HANHCN,stagNet,shu2017cern,ARG,azar2019convolutional,hu2020progressive,pramono2020empowering,yuan2021learningcontext,liu2021multimodal} which is complemented without optical flow. Moreover, we propose an extended version of the framework that utilizes a late fusion strategy as ~\cite{gavrilyuk2020actor,yuan2021learningcontext} to fusion RGB and Flow results. This version reaches the MCA of 94.8\%. Such results illustrate that our framework can achieve competitive performance compared with state-of-the-art methods.

More importantly,
although the performance of our method is slightly lower than~\cite{han2022dualai} in the above experiments, our framework achieves better results under the limited training datasets. To demonstrate this, we conduct experiments on the VD with data ratio of 5\%, 10\%, 25\%, and 50\%. For a fair comparison, we select the same samples as~\cite{han2022dualai}. The results of comparison methods~\cite{PCTDM,ARG,gavrilyuk2020actor,yan2020higcin,ehsanpour2020joint,yuan2021spatio,han2022dualai} are reported directly from~\cite{han2022dualai}. 
\cref{table2} presents the experimental results. 
As can be seen,
our method obviously performs better than the state-of-the-art methods at the data ratio of 5\%, 10\%, 25\%, and 50\%.
Particularly, it achieves the MCA of 79.0\%, 85.6\%, 92.1\%, and 93.2\%, surpassing the existing best results by 2.8\%, 0.1\%, 2.4\%, and 0.5\%, respectively.
Furthermore, by comparing with the baseline model which only consists of the Visual Representation Module, and it predicts the classification score simply from $\mathbf{\overline{X}}$ and $\mathbf{X}$,
our method significantly improves performance with knowledge under limited training data.
In particular, when taking 5\%, 10\%, 25\% samples as the training data, 
our method achieves MCA higher by 12.8\%, 6.8\%, 3.9\% than the baseline model, respectively. 
This clearly demonstrate the effectiveness and superiority of our method under limited training data.

\begin{figure*}[h]
    \centering
    \includegraphics[width=0.97\linewidth]{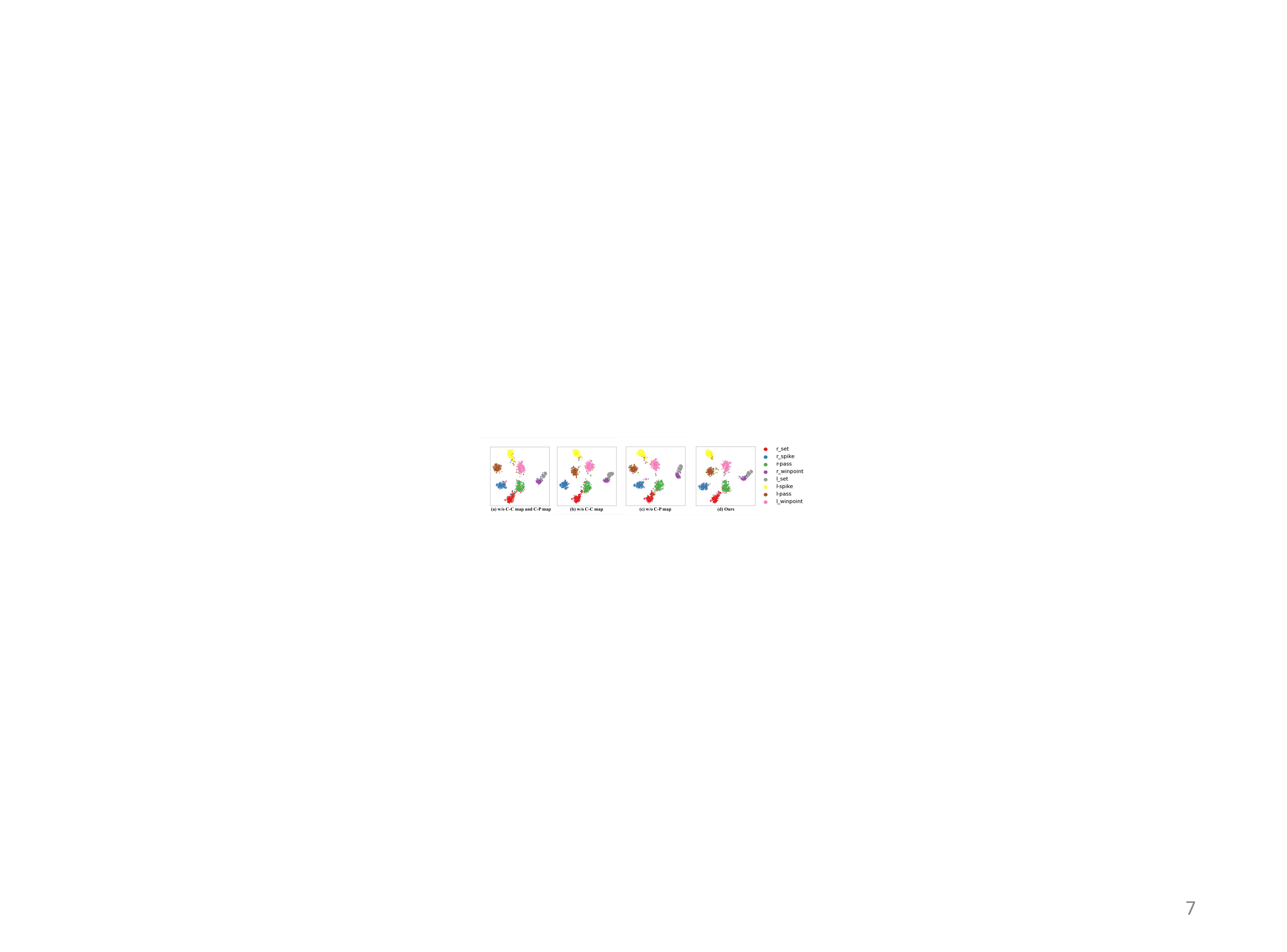}
    \caption{The t-SNE visualization of learned representation by different models on the Volleyball dataset.} 
    \label{fig5}
\end{figure*}

\begin{figure*}[h]
    \centering
    \includegraphics[width=0.97\linewidth]{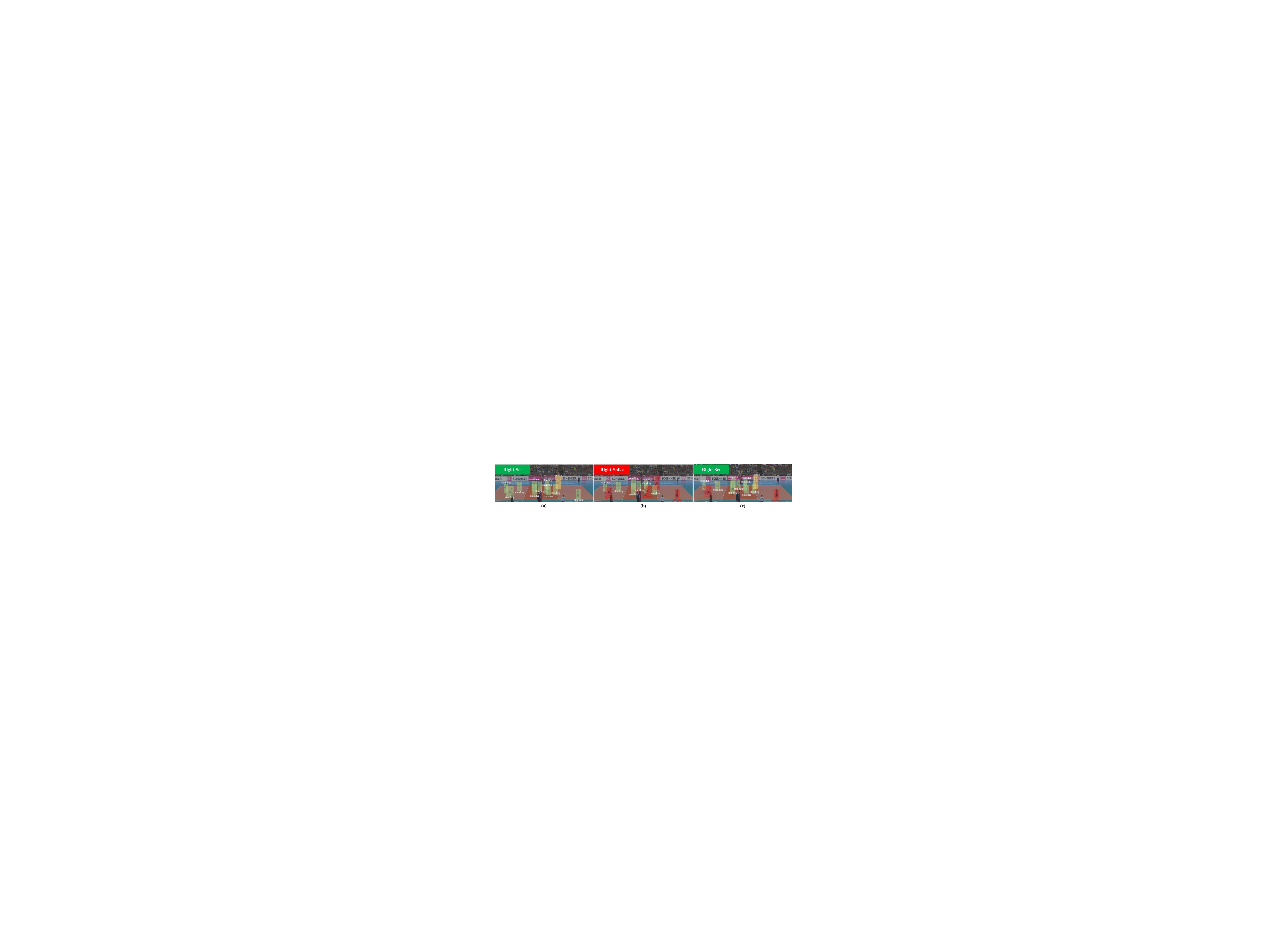}
    \caption{Visualization of the predictions on the Volleyball dataset. (a) The ground truth of individual action labels in \textquotedblleft right-set\textquotedblright~ (b) The model without C-P Map and C-C Map mistakenly classified the activity of \textquotedblleft right-set\textquotedblright~ into the \textquotedblleft right-spike\textquotedblright. (c) Our model with the C-C map and C-P map can classify the activity correctly.} 
    \label{fig6} 
\end{figure*}

\textbf{Results on Collective Activity dataset.}
On CAD, we use both MCA and MPCA for evaluation. Similar to other methods~\cite{azar2019convolutional,yuan2021learningcontext,yuan2021spatio,han2022dualai}, we merge the category ``walking'' and ``crossing'' as ``moving'' to calculate MPCA. In addition, since the scenario of this dataset changes dramatically, there is no certain correlation between action labels and individual positions. 
Therefore, the Class-Position Distribution Map (C-P Map) has not been used in this dataset. 

\cref{table3} reports the experimental results. 
All the compared methods adopt the optical flow.
As can be seen from this table, our method achieves excellent results in terms of the MPCA.
Specifically, our method gains the MPCA of 98.5\% with only RGB input and outperforms all of the compared methods. When using optical flow inputs, it can further improve 0.2\% and achieve the MPCA of 98.7\%.
While for the MCA, our method is slightly lower than the existing best result, this is mainly because “walking” and “crossing” have a high similarity in appearance, which makes our model confused in obtaining C-C Map, and further impact the recognition of these two categories. 

\begin{table}
    \centering
    \caption{Ablation studies of the introduced knowledge on the Volleyball dataset.}
    \begin{tabular}{@{}lc@{}}
    \bottomrule
    Model &  MCA \\
    \midrule
        Base & 93.4\\
        Base + Semantic &  93.6\\
        Base + Semantic + C-P Map & 93.9\\
        Base + Semantic + C-C Map & 94.0\\
    \hline
        \textbf{Base + Semantic + C-C Map + C-P Map} & \textbf{94.5} \\
    \bottomrule
    \end{tabular}
    \label{table4}
\end{table}

\subsection{Ablation Study}
\label{sec4.4}
To investigate the effect of the introduced knowledge in the proposed framework, 
we conduct ablation studies on the VD with the following variants: (A) Base: it only consists of the Visual Representation Module and directly uses the individual representation $\mathbf{\overline{X}}$ and global scene information $\mathbf{X}$ for the final classification. (B) Base + Semantic: it removes both the Class-Class Distribution Map and Class-Position Distribution Map from the overall framework. (C) Base + Semantic + C-P Map: it removes the Class-Class Distribution Map from the whole framework. (D) Base + Semantic + C-C Map: it removes the Class-Position Distribution Map from the whole framework. 

As shown in \cref{table4}, the MCA of model (B) is improved by 0.2\% compared with model (A), which indicates the effectiveness of introducing semantic information. The MCA of model (C) and model (D) is improved by 0.3\%, 0.4\% compared with the result of model (B), respectively. The complete framework introducing both two distribution maps can reach the MCA of 94.5\%, outperforms ablation models by 1.1\%, 0.9\%, 0.6\% and 0.5\%, respectively. These results show the effectiveness of utilizing knowledge to enhance the relation inference process for group activity classification.

\subsection{Visualization}
\textbf{The t-SNE visualization of learned representation.} We adopt t-SNE~\cite{van2008visualizing} to analyze the feature distribution of different models on the VD. As shown in \cref{fig5} (a), feature representations of ``l-spike'' cannot be separated well from ``l-winpoint''. As in \cref{fig5} (b) and (c), when C-P Map or C-C Map is introduced to enhance the relation inference process, our method can distinguish ``l-spike'' and ``l-winpoint'' well. As in \cref{fig5} (d), our final framework is able to differentiate feature representations much better than others. These results obviously demonstrate the effectiveness of introducing knowledge to group activity recognition.

\textbf{The visualization of predictions.} An example of the group activity recognition on the VD is visualized in \cref{fig6}. Compared with ground truth in \cref{fig6} (a), the model without C-P Map and C-C Map mistakenly classified the actions of four players as shown in \cref{fig6} (b), which classified ``setting'' into ``spiking'' may result in the mistake of the group activity recognition. In fact, the appearance of a jumping-like player in \cref{fig6} (b) looks like ``spiking''. Therefore, it is reasonable for such misclassification since the model performs classification mainly based on visual information. In comparison, by introducing the knowledge concretized C-C Map and C-P Map, such as the ``setting'' action appearing in the defensive zone with a higher probability than the ``spiking'' action, our model enhances the visual representation and correctly classifies the ``setting'' action. Therefore, the group activity is correctly classified as shown in \cref{fig6} (c).

\section{Conclusion}
In this paper, we observe that the existing visual representation based group activity recognition methods have not explored the influence of abundant knowledge on the relation modeling process, leading to a limitation of the performance. We propose an idea of knowledge concretization and further present an end-to-end group activity recognition framework. Our framework first utilizes a Visual Representation Module to extract appearance feature, and then a Knowledge Augmented Semantic Relation Module to extract semantic information and explore the semantic relations. Finally, a Knowledge Augmented Semantic Relation Module integrates visual and semantic information with the help of knowledge. Extensive experiments validate that knowledge enable effectively enhance relation inference process and the individual representations. Benefiting from the design of these modules, our framework achieves competitive experimental results on two widely-used datasets.

\bibliographystyle{IEEEtran}
\bibliography{egbib}

\begin{thebibliography}{10}
\providecommand{\url}[1]{#1}
\csname url@samestyle\endcsname
\providecommand{\newblock}{\relax}
\providecommand{\bibinfo}[2]{#2}
\providecommand{\BIBentrySTDinterwordspacing}{\spaceskip=0pt\relax}
\providecommand{\BIBentryALTinterwordstretchfactor}{4}
\providecommand{\BIBentryALTinterwordspacing}{\spaceskip=\fontdimen2\font plus
\BIBentryALTinterwordstretchfactor\fontdimen3\font minus
  \fontdimen4\font\relax}
\providecommand{\BIBforeignlanguage}[2]{{%
\expandafter\ifx\csname l@#1\endcsname\relax
\typeout{** WARNING: IEEEtran.bst: No hyphenation pattern has been}%
\typeout{** loaded for the language `#1'. Using the pattern for}%
\typeout{** the default language instead.}%
\else
\language=\csname l@#1\endcsname
\fi
#2}}
\providecommand{\BIBdecl}{\relax}
\BIBdecl

\bibitem{shi2019skeleton}
L.~Shi, Y.~Zhang, J.~Cheng, and H.~Lu, ``Skeleton-based action recognition with
  directed graph neural networks,'' in \emph{Computer Vision and Pattern
  Recognition}, 2019, pp. 7912--7921.

\bibitem{plizzari2021skeleton}
C.~Plizzari, M.~Cannici, and M.~Matteucci, ``Skeleton-based action recognition
  via spatial and temporal transformer networks,'' \emph{Computer Vision and
  Image Understanding}, vol. 208, p. 103219, 2021.

\bibitem{c3d}
D.~Tran, L.~Bourdev, R.~Fergus, L.~Torresani, and M.~Paluri, ``Learning
  spatiotemporal features with 3d convolutional networks,'' in
  \emph{International Conference on Computer Vision}, 2015, pp. 4489--4497.

\bibitem{i3d}
J.~Carreira and A.~Zisserman, ``Quo vadis, action recognition? a new model and
  the kinetics dataset,'' in \emph{Computer Vision and Pattern Recognition},
  2017, pp. 4724--4733.

\bibitem{ibrahim2016hierarchical}
M.~S. Ibrahim, S.~Muralidharan, Z.~Deng, A.~Vahdat, and G.~Mori, ``A
  hierarchical deep temporal model for group activity recognition,'' in
  \emph{Computer Vision and Pattern Recognition}, 2016, pp. 1971--1980.

\bibitem{bagautdinov2017social}
T.~Bagautdinov, A.~Alahi, F.~Fleuret, P.~Fua, and S.~Savarese, ``Social scene
  understanding: End-to-end multi-person action localization and collective
  activity recognition,'' in \emph{Computer Vision and Pattern Recognition},
  2017, pp. 4315--4324.

\bibitem{ibrahim2018hierarchical}
M.~S. Ibrahim and G.~Mori, ``Hierarchical relational networks for group
  activity recognition and retrieval,'' in \emph{the European conference on
  computer vision}, 2018, pp. 721--736.

\bibitem{GLSTM}
X.~Shu, L.~Zhang, Y.~Sun, and J.~Tang, ``Host–parasite: Graph lstm-in-lstm
  for group activity recognition,'' \emph{IEEE Transactions on Neural Networks
  and Learning Systems}, vol.~32, no.~2, pp. 663--674, 2021.

\bibitem{ARG}
J.~Wu, L.~Wang, L.~Wang, J.~Guo, and G.~Wu, ``Learning actor relation graphs
  for group activity recognition,'' in \emph{Computer Vision and Pattern
  Recognition}, 2019, pp. 9956--9966.

\bibitem{gavrilyuk2020actor}
K.~Gavrilyuk, R.~Sanford, M.~Javan, and C.~G.~M. Snoek, ``Actor-transformers
  for group activity recognition,'' in \emph{Computer Vision and Pattern
  Recognition}, 2020, pp. 836--845.

\bibitem{yuan2021spatio}
H.~Yuan, D.~Ni, and M.~Wang, ``Spatio-temporal dynamic inference network for
  group activity recognition,'' in \emph{International Conference on Computer
  Vision}, 2021, pp. 7476--7485.

\bibitem{li2021groupformer}
S.~Li, Q.~Cao, L.~Liu, K.~Yang, S.~Liu, J.~Hou, and S.~Yi, ``Groupformer: Group
  activity recognition with clustered spatial-temporal transformer,'' in
  \emph{International Conference on Computer Vision}, 2021, pp.
  13\,668--13\,677.

\bibitem{yuan2021learningcontext}
H.~Yuan and D.~Ni, ``Learning visual context for group activity recognition,''
  in \emph{AAAI Conference on Artificial Intelligence}, vol.~35, no.~4, 2021,
  pp. 3261--3269.

\bibitem{wq}
L.~Wu, X.~Lang, Y.~Xiang, Q.~Wang, and M.~Tian, ``Multi-perspective
  representation to part-based graph for group activity recognition,''
  \emph{Sensors}, vol.~22, no.~15, 2022.

\bibitem{han2022dualai}
M.~Han, D.~J. Zhang, Y.~Wang, R.~Yan, L.~Yao, X.~Chang, and Y.~Qiao, ``Dual-ai:
  Dual-path actor interaction learning for group activity recognition,'' in
  \emph{Computer Vision and Pattern Recognition}, 2022, pp. 2990--2999.

\bibitem{liu2021multimodal}
T.~Liu, R.~Zhao, and K.-M. Lam, ``Multimodal-semantic context-aware graph
  neural network for group activity recognition,'' in \emph{2021 IEEE
  International Conference on Multimedia and Expo (ICME)}.\hskip 1em plus 0.5em
  minus 0.4em\relax IEEE, 2021, pp. 1--6.

\bibitem{tang2018mining}
Y.~Tang, Z.~Wang, P.~Li, J.~Lu, M.~Yang, and J.~Zhou, ``Mining
  semantics-preserving attention for group activity recognition,'' in
  \emph{Proceedings of the 26th ACM international conference on Multimedia},
  2018, pp. 1283--1291.

\bibitem{tang2019learning}
Y.~Tang, J.~Lu, Z.~Wang, M.~Yang, and J.~Zhou, ``Learning semantics-preserving
  attention and contextual interaction for group activity recognition,''
  \emph{IEEE Transactions on Image Processing}, vol.~28, no.~10, pp.
  4997--5012, 2019.

\bibitem{contextaware}
A.~Dasgupta, C.~V. Jawahar, and K.~Alahari, ``Context aware group activity
  recognition,'' in \emph{International Conference on Pattern Recognition},
  2021, pp. 10\,098--10\,105.

\bibitem{GINs}
Y.~Tang, Y.~Wei, X.~Yu, J.~Lu, and J.~Zhou, ``Graph interaction networks for
  relation transfer in human activity videos,'' \emph{IEEE Transactions on
  Circuits and Systems for Video Technology}, vol.~30, no.~9, pp. 2872--2886,
  2020.

\bibitem{amer2012cost}
M.~R. Amer, D.~Xie, M.~Zhao, S.~Todorovic, and S.-C. Zhu, ``Cost-sensitive
  top-down/bottom-up inference for multiscale activity recognition,'' in
  \emph{European Conference on Computer Vision}, 2012, pp. 187--200.

\bibitem{lan2012social}
T.~Lan, L.~Sigal, and G.~Mori, ``Social roles in hierarchical models for human
  activity recognition,'' in \emph{Computer Vision and Pattern Recognition},
  2012, pp. 1354--1361.

\bibitem{amer2014hirf}
M.~R. Amer, P.~Lei, and S.~Todorovic, ``Hirf: Hierarchical random field for
  collective activity recognition in videos,'' in \emph{European Conference on
  Computer Vision}, 2014, pp. 572--585.

\bibitem{choi2013understanding}
W.~Choi and S.~Savarese, ``Understanding collective activitiesof people from
  videos,'' \emph{IEEE transactions on pattern analysis and machine
  intelligence}, vol.~36, no.~6, pp. 1242--1257, 2013.

\bibitem{Amer2013}
M.~R. Amer, S.~Todorovic, A.~Fern, and S.-C. Zhu, ``Monte carlo tree search for
  scheduling activity recognition,'' in \emph{International Conference on
  Computer Vision}, 2013, pp. 1353--1360.

\bibitem{wu2021comprehensive}
L.-F. Wu, Q.~Wang, M.~Jian, Y.~Qiao, and B.-X. Zhao, ``A comprehensive review
  of group activity recognition in videos,'' \emph{International Journal of
  Automation and Computing}, vol.~18, no.~3, pp. 334--350, 2021.

\bibitem{wang2017recurrent}
M.~Wang, B.~Ni, and X.~Yang, ``Recurrent modeling of interaction context for
  collective activity recognition,'' in \emph{Computer Vision and Pattern
  Recognition}, 2017, pp. 3048--3056.

\bibitem{stagNet}
M.~Qi, Y.~Wang, J.~Qin, A.~Li, J.~Luo, and L.~Van~Gool, ``stagnet: An attentive
  semantic rnn for group activity and individual action recognition,''
  \emph{IEEE Transactions on Circuits and Systems for Video Technology},
  vol.~30, no.~2, pp. 549--565, 2020.

\bibitem{azar2019convolutional}
S.~M. Azar, M.~G. Atigh, A.~Nickabadi, and A.~Alahi, ``Convolutional relational
  machine for group activity recognition,'' in \emph{Computer Vision and
  Pattern Recognition}, 2019, pp. 7892--7901.

\bibitem{hu2020progressive}
G.~Hu, B.~Cui, Y.~He, and S.~Yu, ``Progressive relation learning for group
  activity recognition,'' in \emph{Computer Vision and Pattern Recognition},
  2020, pp. 980--989.

\bibitem{Detector-Free}
D.~Kim, J.~Lee, M.~Cho, and S.~Kwak, ``Detector-free weakly supervised group
  activity recognition,'' in \emph{Computer Vision and Pattern Recognition},
  2022, pp. 20\,083--20\,093.

\bibitem{shu2017cern}
T.~Shu, S.~Todorovic, and S.-C. Zhu, ``Cern: confidence-energy recurrent
  network for group activity recognition,'' in \emph{Computer Vision and
  Pattern Recognition}, 2017, pp. 5523--5531.

\bibitem{HANHCN}
L.~Kong, J.~Qin, D.~Huang, Y.~Wang, and L.~Van~Gool, ``Hierarchical attention
  and context modeling for group activity recognition,'' in \emph{2018 IEEE
  International Conference on Acoustics, Speech and Signal Processing}, 2018,
  pp. 1328--1332.

\bibitem{CCG}
J.~Tang, X.~Shu, R.~Yan, and L.~Zhang, ``Coherence constrained graph lstm for
  group activity recognition,'' \emph{IEEE Transactions on Pattern Analysis and
  Machine Intelligence}, pp. 1--1, 2019.

\bibitem{yan2018participation}
R.~Yan, J.~Tang, X.~Shu, Z.~Li, and Q.~Tian, ``Participation-contributed
  temporal dynamic model for group activity recognition,'' in \emph{Proceedings
  of the 26th ACM international conference on Multimedia}, 2018, pp.
  1292--1300.

\bibitem{PCTDM}
------, ``Participation-contributed temporal dynamic model for group activity
  recognition,'' in \emph{ACM international conference on Multimedia}, 2018,
  pp. 1292--1300.

\bibitem{yan2020higcin}
R.~Yan, L.~Xie, J.~Tang, X.~Shu, and Q.~Tian, ``Higcin: hierarchical
  graph-based cross inference network for group activity recognition,''
  \emph{IEEE Transactions on Pattern Analysis and Machine Intelligence}, 2020.

\bibitem{lu2019gaim}
L.~Lu, Y.~Lu, R.~Yu, H.~Di, L.~Zhang, and S.~Wang, ``Gaim: Graph attention
  interaction model for collective activity recognition,'' \emph{IEEE
  Transactions on Multimedia}, vol.~22, no.~2, pp. 524--539, 2019.

\bibitem{ehsanpour2020joint}
M.~Ehsanpour, A.~Abedin, F.~Saleh, J.~Shi, I.~Reid, and H.~Rezatofighi, ``Joint
  learning of social groups, individuals action and sub-group activities in
  videos,'' in \emph{European Conference on Computer Vision}, 2020, pp.
  177--195.

\bibitem{sbgar}
X.~Li and M.~C. Chuah, ``Sbgar: Semantics based group activity recognition,''
  in \emph{International Conference on Computer Vision}, 2017, pp. 2895--2904.

\bibitem{kay2017kinetics}
W.~Kay, J.~Carreira, K.~Simonyan, B.~Zhang, C.~Hillier, S.~Vijayanarasimhan,
  F.~Viola, T.~Green, T.~Back, P.~Natsev \emph{et~al.}, ``The kinetics human
  action video dataset,'' \emph{arXiv preprint arXiv:1705.06950}, 2017.

\bibitem{detr}
N.~Carion, F.~Massa, G.~Synnaeve, N.~Usunier, A.~Kirillov, and S.~Zagoruyko,
  ``End-to-end object detection with transformers,'' in \emph{European
  Conference on Computer Vision}, 2020, pp. 213--229.

\bibitem{wang2021end}
Y.~Wang, Z.~Xu, X.~Wang, C.~Shen, B.~Cheng, H.~Shen, and H.~Xia, ``End-to-end
  video instance segmentation with transformers,'' in \emph{Computer Vision and
  Pattern Recognition}, 2021, pp. 8741--8750.

\bibitem{he2017mask}
K.~He, G.~Gkioxari, P.~Doll{\'a}r, and R.~Girshick, ``Mask r-cnn,'' in
  \emph{International Conference on Computer Vision}, 2017, pp. 2961--2969.

\bibitem{nair2010rectified}
V.~Nair and G.~E. Hinton, ``Rectified linear units improve restricted boltzmann
  machines,'' in \emph{ICML}, 2010.

\bibitem{fu2019dual}
J.~Fu, J.~Liu, H.~Tian, Y.~Li, Y.~Bao, Z.~Fang, and H.~Lu, ``Dual attention
  network for scene segmentation,'' in \emph{Computer Vision and Pattern
  Recognition}, 2019, pp. 3146--3154.

\bibitem{pramono2020empowering}
R.~R.~A. Pramono, Y.~T. Chen, and W.~H. Fang, ``Empowering relational network
  by self-attention augmented conditional random fields for group activity
  recognition,'' in \emph{European Conference on Computer Vision}, 2020, pp.
  71--90.

\bibitem{van2008visualizing}
L.~Van~der Maaten and G.~Hinton, ``Visualizing data using t-sne.''
  \emph{Journal of machine learning research}, vol.~9, no.~11, 2008.

\end{thebibliography}

\end{document}